# Adaptive Feature Processing for Robust Human Activity Recognition on a Novel Multi-Modal Dataset

Mirco Möncks, Varuna De Silva, Jamie Roche, and Ahmet Kondoz

*Abstract*— Human Activity Recognition (HAR) is a key building block of many emerging applications such as intelligent mobility, ambient-assisted living and human-robot interaction. With robust HAR, systems will become more human-aware, leading towards much safer and empathetic autonomous systems. While human pose detection has made significant progress with the dawn of deep convolutional neural networks (CNNs), the state-of-the-art research has almost exclusively focused on a single sensing modality, especially video. However, in safety critical applications it is imperative to utilize multiple sensor modalities for robust operation. To exploit the benefits of state-of-the-art machine learning techniques for HAR, it is extremely important to have multimodal datasets. In this paper, we present a novel, multi-modal sensor dataset that encompasses nine indoor activities, performed by 16 participants, and captured by four types of sensors that are commonly used in indoor applications and autonomous vehicles. This multimodal dataset is the first of its kind to be made openly available and can be exploited for many applications that require HAR, including sports analytics, healthcare assistance and indoor intelligent mobility. We propose a novel data preprocessing algorithm to enable context dependent feature extraction from the dataset to be utilized by different machine learning algorithms. Through rigorous experimental evaluations, this paper reviews the performance of machine learning approaches to posture recognition, and analyses the robustness of the algorithms. When performing HAR with the RGB-Depth data from our new dataset, machine learning algorithms such as a deep neural network reached a mean accuracy of up to 96.8% for classification across all stationary and dynamic activities.

*Index Terms*— Human Activity Recognition, HAR, Machine Learning, Autonomous Driving, Pose Estimation, Indoor Mobility, LboroHAR dataset

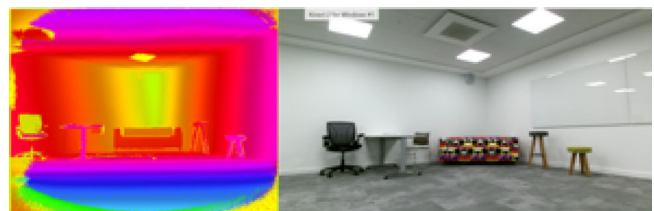

(a) Depth sensor and RGB color image (RGB-D)

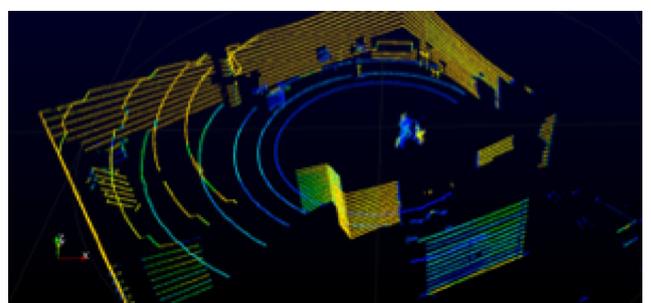

(b) LiDAR Sensor

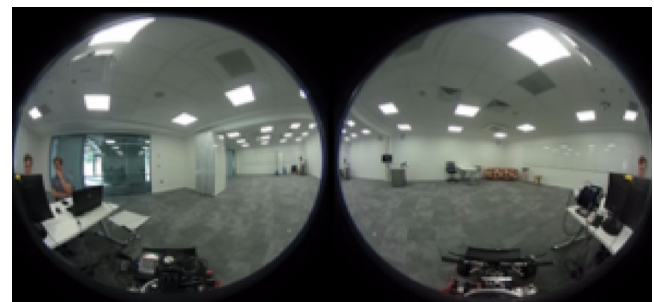

(c) RGB 360º camera

Fig. 1. Example of sensor modalities used in LboroHAR Dataset

## I. INTRODUCTION

Recent developments within the fields of artificial intelligence (AI), machine learning (ML), advances in cyber-physical systems, and human-machine interaction allow technology to become more human-centric. This eventually enables machines to support people in their daily lives and increase the effectiveness of human work force by supporting harmful or physically demanding tasks. By allowing machines to intervene in human decision-making processes, it must be ensured that they effectively respond to human influence and interaction. Towards this end, machines need to analyze sensor data streams to detect, recognize, and precisely predict human activities. This branch of applied machine learning is known as Human Activity Recognition (HAR). HAR is an important element of many emerging applications such as personal assistant robots, retail analytics, sports analytics, smart homes and in connected autonomous vehicles (CAVs) [1]–[7]. For example, HAR enabled systems can be applied to check the rate of degeneration of neurological disorders [8]. In certain applications like in CAVs, HAR acts as a safety critical block, where, incorrect or inadequate pedestrian detection could lead to fatal accidents, thus indicating the ongoing need for further improvement to HAR [9].

The majority of the research in the HAR is originated from

Mirco Möncks, Jamie Roche, Varuna De Silva, Ahmet Kondoz are with Institute for Digital Technologies, Loughborough University London, 3 Lesney Avenue, The Broadcast Centre, Here East, Queen Elizabeth Olympic Park, London, E15 2GZ, United Kingdom, e-mail: mirco.moencks@gmail.com e-mail: A.J.Roche@lboro.ac.uk, e-mail: V.D.De- Silva@lboro.ac.uk, e-mail: A.Kondoz@lboro.ac.uk
Varuna De Silva is the corresponding author.

computer vision literature on analyzing videos/images, and since the recent developments in Internet of Things other sensors such as body worn accelerometers has been used for HAR. The computer vision based HAR systems can be effectively seen as two components: body pose estimation and the temporal classification of body pose changes [10]–[12]. As such, body pose estimation, which predicts the human body joints on a video frame, acts as a feature extraction step. Body pose estimation has massively benefitted from recent advances in convolutional neural networks (CNNs) [10], [11], [13]. It is often mentioned that effective feature extraction is the major challenge to robust HAR [12], [13]. Recent work, therefore, strives to increase the information density of respective feature vectors without increasing its dimensionality [14]. Due to the curse of dimensionality [15], and increased computation efforts, exploiting all extracted or available features often does not result in higher classification accuracy. The method of static / inelastic feature selection is particularly well suited to classifications that differ explicitly. In the static feature selection method, the subset of features for a feature vector is predefined manually (or automatically by other ML algorithms) and not altered during the classification task. For example, the classification of the activities "walking" and "sitting" can be realized by selecting the velocity of participants' center of mass. However, the accurate classification of activities that differ less explicitly requires case related consideration of various features. For example, considering the velocity of hip movement in sitting activities is of minor interest. On the other hand, for the classification of dynamic activities such as walking, jogging and sprinting, it is of particular interest to include the relative velocity of the joints of the lower body into a feature vector. Consequently, the importance of dynamic or elastic feature selection for enhanced classification accuracy is a major interest in previous work [16]–[18].

The recent fatal accident involving a driverless vehicle and a pedestrian accentuates the fact that a single sensor stream cannot be relied upon for HAR. Therefore, utilizing multiple sensor streams that offer sensing diversity, i.e. multimodality, is crucial for the success of HAR activities in safety critical applications such as CAVs. Therefore, multimodal data analysis is an important avenue that warrants further research towards the success of HAR. Availability of appropriate multimodal datasets is crucial for this purpose. For mobility solutions in outdoor environments, previous work mostly relied on multi- and omni-directional cameras and laser sensors [19]–[22]. Research related to indoor HAR often exploits RGB-D sensors, stereo video camera or wearable sensors [1], [23]–[25]. However, none of the multimodal datasets reviewed encompass sensor data that are commonly used for both indoor and outdoor HAR. Compared with outdoor environments such as driverless cars, there is a lack of rigorous research experimentally testing multimodal HAR algorithms for indoor mobility applications (IMA). This is critical because humans tend to have a different behavior and attention spans in indoor environments [26].

In order to enable robust HAR in applications that work both indoors and outdoors, this paper presents a novel algorithm for elastic (dynamic) feature processing method and new multimodal dataset for use by the research community. The contributions of this paper are as follows:

- After a thorough investigation of the limitations of existing work, we present a novel multi-modal sensor dataset for indoor HAR of AV (Different sensor modalities used are illustrated in Figure 1) and
- We propose a novel data pre-processing and flexible feature recognition technique to improve classifier performance of state-of-the-art machine learning techniques.

This paper is organized as follows: section 2 provides a literature review on the applicability of HAR in its various contexts and scrutinizes both the underlying sensors and algorithms. Section 3 describes our methodology, followed by the presentation of our results in section 4. Whilst concluding the paper in section 5, we state that under dynamic conditions the accuracy of the classification approaches is inadequate. For future work, we suggest fusing RGB-D and LiDAR data to overcome major limitations of both sensors.

## II. RELATED WORK

In this section, we analyze the application range of HAR. We also discuss state-of-the- art datasets for HAR with a focus on intelligent mobility applications. We provide evidence that there is no sensor dataset that can sufficiently address the gap of HAR for Intelligent Mobility Applications.

### A. Healthcare and Ambient-Assisted Living

HAR is often used in ambient-assisted living environments, hospitals or rehabilitation clinics; especially for elderly care. Here it is of interest to identify scenarios which require urgent medical assistance, thus differentiating daily-life activities from abnormal or potentially harmful behavior [27]–[29]. Since falling implies one of the biggest risks to the health of elderly people [30], this activity is addressed by many previous works [31]–[33]. Other research also strives to recognize vomiting, chest pain, and fainting [2]. Work that is more related to prophylactic healthcare analyses people's movements and calculates calories burnt during the day [34]. Moreover, HAR is commonly used as a diagnostic tool to check the rate of degeneration of neurological disorders [8]. Recent work is aiming towards more proactive approaches in ambient- assistant living where a system responds to the anticipated intention of people, rather than the actual activity itself [35].

### B. Smart Mobility and Autonomous Vehicles

Efforts which are more related to vehicles strive to decrease road accidents by recognizing human traffic participants, establishing a communication system between vehicles and smart devices [36], or reporting accidents in a timely manner [37]. In this context, [38] or [39] provide a promising approach for pedestrian intention recognition: a state-of-the- art sensor system, which matches the predicted intention of a pedestrian with the current direction of a driver's attention, and initiates an emergency brake when needed. This research is related to scenarios with vehicles driven by humans in outdoor environments. Among others, a robust model for autonomous vehicles predicting pedestrian's intention in indoor environments remains open [40].

TABLE I
COMPARISON OF HAR DATASETS

| Reference | Classes | Subjects | Sensors |
|---|---|---|---|
| [47] | 20 | 10 | RGB-D |
| [48] | 27 | 8 | RGB-D (and wearable sensors) |
| [49] | 30 | 10 | RGB-D |
| [45] | 60 | 40 | RGB-D |
| [50] | 10 | 10 | Wearable RGB-D |
| [51] | n. a. | n. a. | Accelerometer |
| [52] | 10 | 11 | Wearable (Google Glasses) |
| [53] | 12 | 49 | Wearable stereo video camera |
| [54] | 65 | n. a. | Stereo video camera |
| [55] | 52 | 10 | Stereo video camera |
| [46] | n.a. | 12 | Stereo video camera |
| [11] | 473 | 40k | Stereo video camera (2D Images) |

## C. Human Activity Recognition Datasets

Derived from notable surveys addressing 3D data for HAR [4], [41]–[43], and recent work related to camera analysis and wearable technologies [44]. Table I provides an overview about the datasets that could be considered useful for addressing our research gap. It can be seen that there are three different classes of datasets in the literature: (a) stereo images captured by video cameras (RGB), (b) three-dimensional (3D) images captured by depth sensors (RGB-D), and (c) data captured by wearable sensors (e.g. acceleration of specific body joints). In terms of subjects, samples and classes, the NTU RGB-D dataset [45] represents one of the most comprehensive datasets for HAR. The activities performed in [36] or [46] deviate significantly from our focus. The MPII Human Pose dataset can be considered a state of the art benchmark for the evaluation of articulated 2D human pose estimation [11], [12]. For navigation tasks, however, depth information as well as sequential movements and related velocities of body joints are of necessity. Consequently, this dataset is not suitable for our research objective. With the data captured, the dataset in [47] points into the right direction of our research. Due to the use of outdated sensors, this dataset would not be longer adequate for our research intention.

## D. Datasets for CAV Research suitable for HAR

In the context of autonomous driving, we identified five considerable datasets using various sensors. The CamVid Database was one of the first experimentally collected video dataset with object class semantic labels for the purpose of visual object analysis [20]. Captured from the perspective of a driving vehicle and semantically labelled, these images address the need for experimental data to quantitatively evaluate emerging algorithms. The work in [19] presents a comprehensive dataset collected with a multimodal sensor ensemble attached to an autonomous ground vehicle testbed. We also identified work which is more related to real urban street scenarios, suited for the appearance-based recognition methods [21], [56]. The combination of the sensors is ideal for navigation, localization or mapping, but at the same time limits the implementation of HAR considerably. A high number of pedestrians and the respective 3D-depth information can be found in [22]. The autonomous driving datasets available are primarily addressing the challenges of (a) urban scene under- standing, (b) state-of-the-art computer vision, (c) simultaneous localization and mapping for AV. They mostly rely on sensors such as lidar and video camera. The gathered data is optimized to detect objects and people in traffic.

TABLE II
LBOROHAR DATASET: OVERVIEW.

| Name | Description |
|---|---|
| Title | LboroLdnHAR |
| Summary | human activities captured by three sensors |
| Date of Collection | 17/06/2018; 18/06/2018 |
| Location | Loughborough University London - London, United Kingdom |
| Size | 16 participants x 9 activities x 3 Sensors (RGB-D; Lidar; 360° Camera) |
| Activity Classes | {1 - sitting on office chair; 2 - standing and texting; 3 - sitting on stool; 4 - lying on couch; 5 - walking; 6 - walking and texting; 7 - carrying objects; 8 - pulling object; 9 - running} |
| Data Streams | 360° Camera Stream (each file contains a scenario, ca. 2 min); Lidar Stream (each file contains a scenario, ca. 2 min); RGB-D Stream (captured by RGB-D sensor); |
| Extracted Parameters for further Consideration | coordinates (51 frames of respective activity); velocity (50 frames of respective activity); acceleration (49 frames of respective activity) |
| Extracted Body Joints for further Consideration | Head; Neck; Chest; MiddleSpine; LowerSpine; Hip; Center_of_mass; Center_of_mass_projection_to_ground; LHand; REye; EffectorHead; RClavicle; RShoulder; RForearm; RHand; LClavicle; LShoulder; LForehand; LHand; RThigh; RShin; RFoot; RToe; EffectorRToe; LThigh; LShin; LFoot; LToe; EffectorToe |
| Orientation | Cartesian Coordinate System |

## E. Motivations for the proposed work

It is often mentioned that effective feature extraction is a major challenge to HAR [12]. Most recent work proposes to tackle this issue by increasing the information density of respective feature vectors without increasing its dimensionality. While most literature has focused on elastic (dynamic) feature selection methods, we propose a static feature selection method as a feature processing technique.

Our review illustrates that the research in most activity classes captured were derived from the outdoor datasets reviewed. Research related to indoor HAR often applies RGB-D sensors, stereo video camera or wearable sensors. For outdoor environments, most HAR systems rely on multi- and omni-directional cameras and laser sensors. The available datasets do not encompass sensors that are commonly used for both indoor and outdoor HAR. In other words, there is a research gap in HAR for AV considering hybrid (indoor-outdoor) scenarios. This gap is critical in so far as ambient-assistant living assistants need to be able to support people in every situation of everyday life. Such an assistant must be able to navigate the way in outdoor environments and both indoor environments. We could not identify any previous work that considers this aspect.

## III. THE PROPOSED ALGORITHM AND THE DATASET

To overcome the limitations summarized in the previous section, we created a dataset with 16 participants, performing 9 activities. The activities were captured by a LiDAR, RGB-

**Panel (a):** activity 1 - Sitting on an office chair (camera and IR sensor perspective with evacuated background).

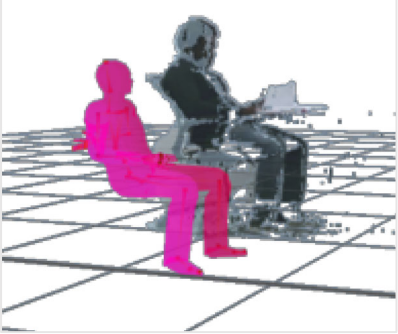

**Panel (b):** activity 2 - Standing and texting with smart phone: retro fitted 3D representation of participant.

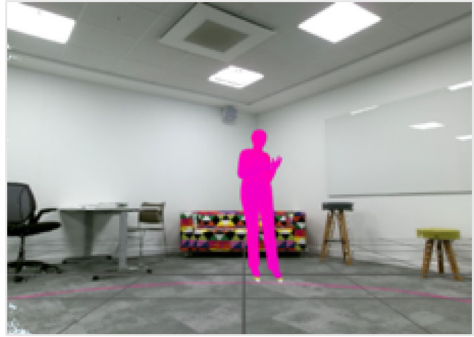

**Panel (c):** activity 3 - Sitting on a stool: camera and IR sensor perspective with evacuated background.

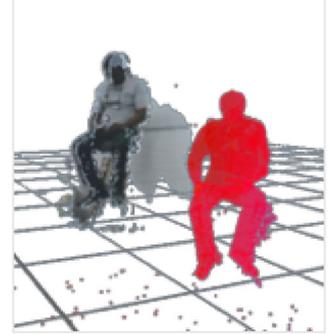

**Panel (d):** activity 4 - laying on a couch: retro fitted 3D representation of participant.

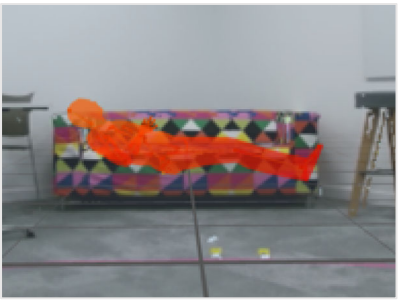

**Panel (e):** activity 5: Walking - camera and IR sensor perspective with evacuated background.

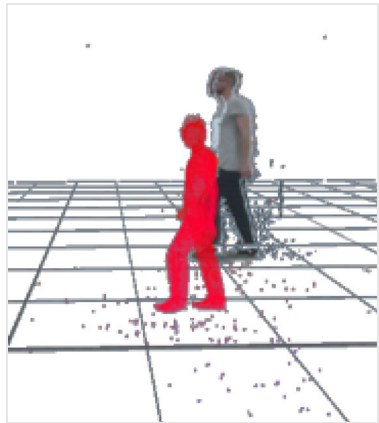

**Panel (f):** activity 6 - Walking while texting: camera and IR sensor perspective with evacuated background.

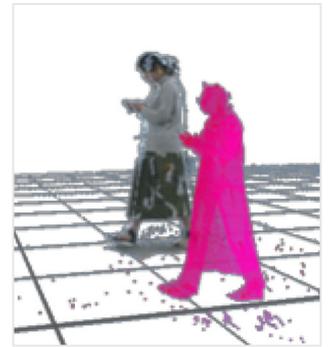

**Panel (g):** activity 7 - Carrying Boxes: camera and IR sensor perspective with evacuated background.

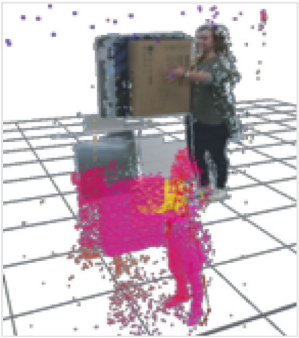

**Panel (h):** activity 8 - Moving Object: camera and IR sensor perspective with evacuated background.

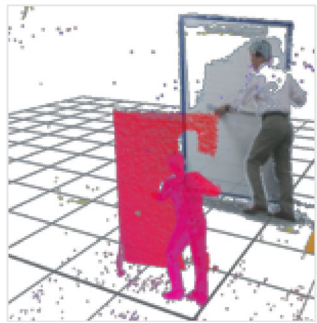

**Panel (i):** activity 9 - Running: camera and IR sensor perspective with evacuated background.

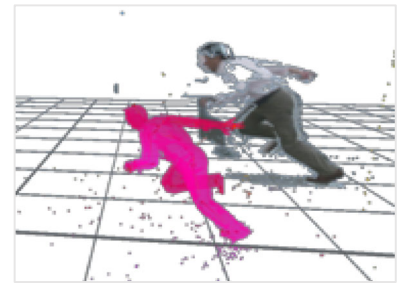

Fig. 2. LboroHAR dataset: Activities.

D, and 360°- camera. Furthermore, we propose a novel algorithm allowing flexible feature extraction and data utilization which improves the performance of multiple machine learning algorithms.

*A. Data Collection*

Our LboroHAR dataset contains data of typical human activities in indoor environments, captured by three sensors (Table II). The 9 activities performed by 16 participants were chosen on the basis of related work (Figure 2). To enable a comparison of the LboroHAR dataset with other datasets, most activity classes captured were derived from the datasets reviewed (Figure 1, Figure 2). We focused on those indoor activities where humans have a limited attention span, resulting in scenarios where it is highly likely that humans do not pay attention to an indoor AV. In addition to our research, this approach facilitates the later derivation of driving policies for indoor AV in critical scenarios.

For further analysis, we solely considered the most distinguishing postures of each participant for each activity. When the coordinates of respective body joints were of interest, we selected 51 frames per person and activity (50 fames for velocity, and 49 frames for acceleration respectively). Our publicly made LbroHAR dataset, however, contains the whole sequence of activities. This allows a wider usability of our work.

*B. DAEFE – Data Pre-processing Algorithm for Elastic Feature Extraction*

We propose an elastic feature extraction method to overcome shortcomings of previous algorithms that solely allow inelastic feature extraction for the utilization of one particular machine learning algorithm [57], [58]. Our algorithm encompasses three main steps. They are represented in Figure 3, and are discussed in the following:

1) Parameter of Interest: The algorithm allows the selection of specific body joints of interests, and their describing parameters of interest (coordinates, velocity, or acceleration). We intentionally refrain from automatic feature extraction with CNN to enable the addressing of potential research interest that are related to specific body joints, or regions of the human body.
2) Feature extraction: The coordinates (velocities or accelerations) of the skeleton joints are used to create the respective feature vectors which represent human postures.
3) Activity Features Computation: A matrix containing all feature vectors that represent the activities is created and used for classification. For supervised ML algorithms, this matrix is extended by one row labelling the activity.

Instead of using all frames captured, we solely consider a subset of $P$ frames that depict significant poses per action and participant, to increase generalizability and decrease complexity:

$$P \in \{51, 50, 49\}. \quad (1)$$

Then, the features of interest need to be extracted from skeleton data representing the input to our ML algorithms. This can be done by an in-built ready to use joint extraction algorithm included in the RGB-D libraries [57], [59]. The selection of these features is motivated by the fact that it is applied in related work, thus allowing us to evaluate the performance of the algorithms in comparison to previous work reviewed. Moreover, the joint extraction algorithm allows us a compact representation of the human body; extracted body joints can then be used as features [3], [4]. A straight-forward approach for the feature extraction can be seen in considering locations of body joints, distances, velocity or acceleration, while relying on spatial information. Previous work, however, typically relies on more complex features that are based on the estimation of a plane that intersects with some joints [57]. The features can then be extracted by measuring the distances between plane and body joints.

Following [58], our algorithm also utilizes spatial features computed from 3D skeleton data (Figure 4). The neglection of the respective time stamp eliminates temporal dependency. In order to evaluate how HAR can be performed by considering the speed of movement, we integrate velocity and acceleration of body joints. Each joint is represented by a 3D vector in the cartesian coordinate space of the RGB-D sensor $J_i$. Specific activities were performed in certain regions of the test area. For instance, participants always stand on the same place while texting (class 2). It is, therefore, necessary to make the activity independent of the region in which it was carried out. To compensate this effect, we normalized the data by replacing the absolute coordinate system with a relative coordinate system centered on one body joint. Following [49], we consider a skeleton of $n$ body joints, $J_0$ being the coordinates of the head joint, $J_1$ being the coordinates of the neck joint, the $i$-th joint feature $f_i$ is the distance vector between $J_0$ and $J_1$:

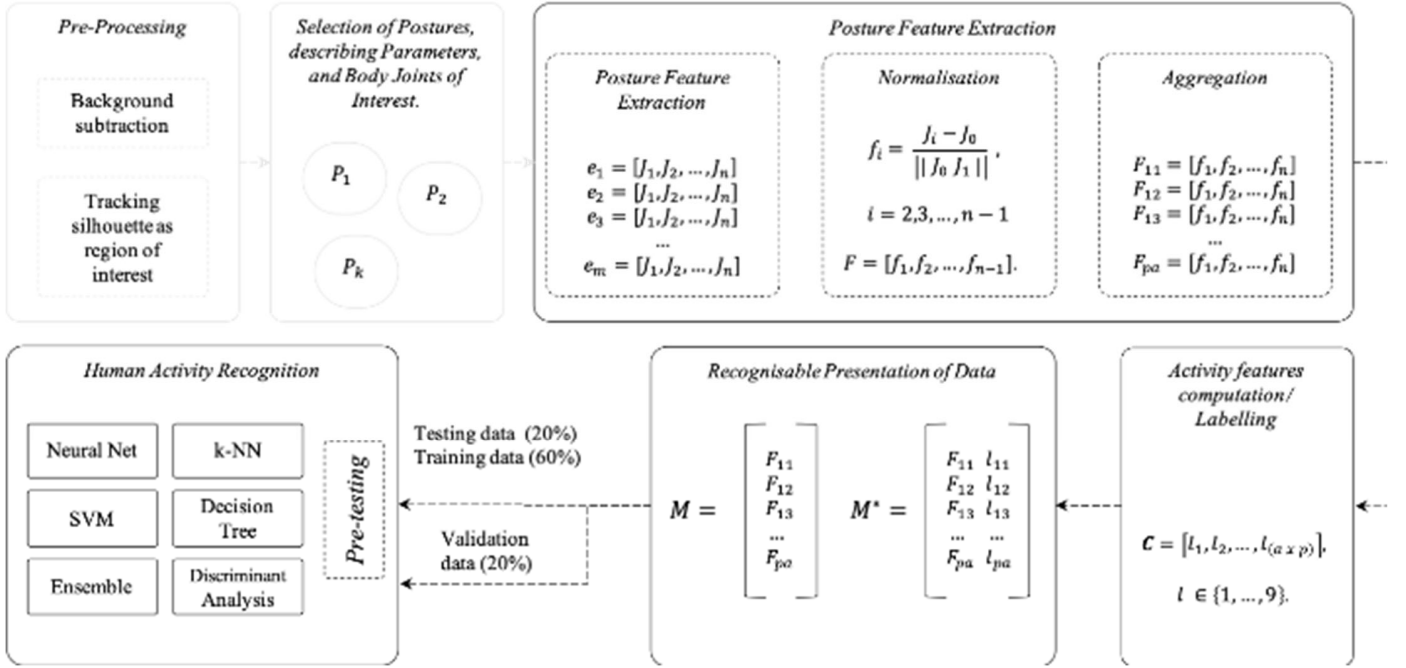

Fig. 3. DAEFE – Data Pre-processing Algorithm for Adaptive (Elastic) Feature Extraction

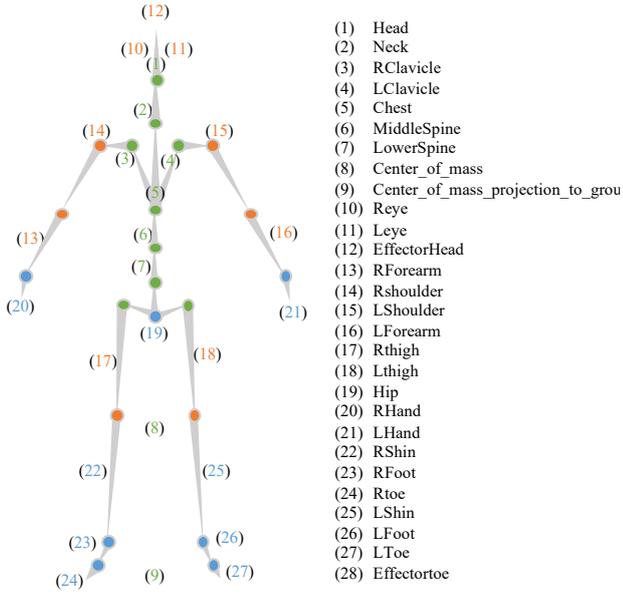

Fig. 4. Subsets of joints considered in the evaluation of the selected ML algorithms

(1) Head
(2) Neck
(3) RClavicle
(4) LClavicle
(5) Chest
(6) MiddleSpine
(7) LowerSpine
(8) Center_of_mass
(9) Center_of_mass_projection_to_grou
(10) Reye
(11) Leye
(12) EffectorHead
(13) RForearm
(14) Rshoulder
(15) LShoulder
(16) LForearm
(17) Rthigh
(18) Lthigh
(19) Hip
(20) RHand
(21) LHand
(22) RShin
(23) RFoot
(24) Rtoe
(25) LShin
(26) LFoot
(27) LToe
(28) Effectortoe

$$f_i = \frac{J_i - J_0}{||J_0\, J_1||}, \qquad i = 2, 3, \dots, n-1 \qquad (2)$$

These features are invariant to the position of the participant within the coverage area of the RGB-D sensor. They may be seen as a set of distance vectors which connect each joint to the joint of the torso. A posture feature vector $F$ can then be created for each frame captured:

$$F = [f_1, f_2, \dots, f_{n-1}]. \qquad (3)$$

Penultimately, we aggregate each feature vector $F$ of activity $a$ and participant $p$ to a combined input matrix $M$ which is exploited by the algorithm:

$$M = \sum_{p=1}^{16} \sum_{a=1}^{9} [F_{p,a}]. \qquad (4)$$

For the purpose of testing unsupervised learning algorithms, we lastly created a classification vector $C$, that encompasses the label of the activity described by the respective feature vectors:

$$C = [l_1, l_2, \dots, l_{(a \times p)}], \qquad l \in \{1, \dots, 9\} \qquad (5)$$

$$M^* = [M, C^T]. \qquad (6)$$

The matrix M (and $M^*$ for supervised learning procedures respectively) represents the data collected in an organized, recognisable form that can now be processed by ML algorithms. This work follows the common approach and splits the dataset into three disjointed entities for the purpose of training (60%), testing (20%), and validation (20%). After training a respective model, the validation dataset is used as a proxy for generalization error underlying. S-fold cross-validation (S = 5) was applied, in order to provide a sufficient number of data points to refine the models [62].

The selection of six algorithms is based on initial pre-tests (Table III). We evaluated the performance of 22 ML algorithms for their applicability and continued our experiments only with those which performed best against similar algorithmic types: fine decision tree, bagged trees, fine k-nearest neighbor classifier, cubic support vector machine, and deep neural networks.

TABLE III
LBOROHAR DATASET: OVERVIEW.

| Type | Name | Accuracy |
|---|---|---|
| Decision Tree | Fine Tree | 79.20% |
| Decision Tree | Medium Tree | 59.90% |
| Decision Tree | Coarse Tree | 45.40% |
| Discriminant Analysis | Quadratic Discriminant | Failed |
| Discriminant Analysis | Linear Discriminant | 80.00% |
| Ensemble | Bagged Trees | 92.50% |
| Ensemble | Subspace k-NN | 89.50% |
| Ensemble | Subspace Discriminant | 78.60% |
| Ensemble | Boosted Trees | 63.70% |
| Ensemble | EUSBoosted Trees | 59.90% |
| Neural Network | Deep Neural Net | 95.00% |
| k-NN | Fine k-NN | 94.40% |
| k-NN | Weighted k-NN | 91.80% |
| k-NN | Cosine k-NN | 87.40% |
| k-NN | Cubic k-NN | 86.20% |
| k-NN | Medium k-NN | 85.40% |
| SVM | Cubic SVM | 97.90% |
| SVM | Quadratic SVM | 92.80% |
| SVM | Fine Gaussian SVM | 91.60% |
| SVM | Linear SVM | 81.90% |
| SVM | Medium Gaussian SVM | 79.40% |
| SVM | Coarse Gaussian SVM | 65.10% |

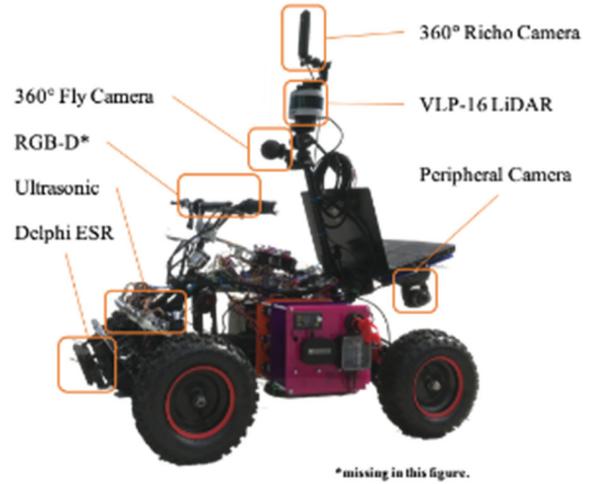

Fig. 5. Loughborough University London Autonomous Driving Sensor Test Bed.

## IV. EXPERIMENTAL RESULTS AND DISCUSSIONS

### A. Experimental Setup

#### 1) Sensor Test Bed

To draw conclusions from our experimental research regarding generalizability and applicability for HAR of AV in indoor environments later on, we endeavored to create the most realistic conditions possible. Consequently, we used an autonomous driving testbed developed by [63], which is shown in Figure 5. The LboroHAR dataset encompasses (raw and pre-processed) data from three sensors attached to this vehicle: (a) IR RGB-D Sensor (Sensor 3), (b) VLP-16 LiDAR (Sensor 4), and (c) 360° Richo RGB-Camera (Sensor 5). These sensors, or the sensor types, are commonly used in the automotive industry for the purpose of assisted or autonomous driving [19], [21].

*2) Environmental Setup*

To evaluate the general applicability of ML algorithms for certain purposes – or to compare the effectiveness of various sensors accordingly – it is critical to guarantee an experimental setup that allows optimal parameter input for subsequent tests [64]. In other words, we need to provide conditions where each sensor reaches its optimal performance level by reducing the impact of sensor-specific limitations, such as light conditions or minimum operative distance, to a negligible amount. This premise eventually led to our experimental setup depicted in Figure 1.

## V. RESULTS

We have laid the foundation for fully understanding our methodology and enabling future work to reproduce our experiment. Therefore, we now present our experimental results that answer our research intention. When similar patterns occur in each algorithm, we exemplify these findings and results by referring to one specific ML approach, rather than individually mentioning all algorithms considered.

### A. Applicable Feature Modalities

We tested which of the extracted feature modalities, namely relative coordinates $c_J$, velocity $v_J$ and acceleration $a_J$ of body joints, are applicable for the purpose of HAR. As presumed, selected algorithms perform best when using $c_J$ as input parameter (Table IV). The accuracy of correctly predicted classes significantly drops when using $v_J$, or $a_J$ accordingly. A reason for this can be found in our initial considerations and the algorithmic classes used. We are striving to classify human activities based on time-invariant spatial posture recognition. Although the amount of existing data of our feature vector remains constant, the information density decreases continuously. This is mainly due to the physical relationship of the measured quantities. The use of $c_J$ allows pattern recognition in different postures. However, especially with static postures, $v_J$ and $a_J$ of individual body joints do not provide any information about the underlying activity.

TABLE IV
TEST RESULTS OF APPLICABLE FEATURES FOR HAR

| Classifier | Overall Accuracy ($c_J$) | Overall Accuracy ($v_J$) | Overall Accuracy ($a_J$) |
|---|---|---|---|
| Decision Tree | 79.20% | 34.20% | 17.00% |
| Lin. Discriminant. | 80.00% | 29.40% | 12.90% |
| Cubic SVM | 91.60% | 44.00% | 24.40% |
| Fine k-NN | 94.40% | 45.50% | 21.60% |
| Bagged Tree | 92.50% | 46.30% | 22.50% |
| DNN | 95.10% | 29.00% | 10.80% |

Fig. 7. Aggregated Confusion Matrix of k-NN Classifier

Fig. 8. Confusion Matrix of Deep Neural Network Classifier

This may lead to a first conclusion that coordinates are the only suitable input parameters that should be used for reliable classification. On a closer look, however, there are some issues with such a statement. Velocity and acceleration tend to be insufficient indicators to classify all types of human activity. However, classification of dynamic activities (5-9) is comparatively accurate under these conditions (Figure 8). The overall objective of this research field is to derive driving policies for AV and enable safe human-robot interaction. Therefore, it is rather necessary to estimate future activities after classifying human beings and its current activity. For example, determining the future location of a pedestrian crossing a street requires consideration of its velocity. Most of the related work that addresses HAR with visual sensors relies on a feature space of coordinates. To ensure comparability we therefore proceed with $c_J$ as applicable input variables.

TABLE V
RELATIONSHIP OF FEATURES AND CLASSIFICATION ACCURACY

| Classifier | PCA enabled (Y/N) | Accuracy ($c_9$ \| $c_{J,x}$, $c_{J,y}$, $c_{J,z}$) | Accuracy ($c_9$ \| $c_{J,x}$, $c_{J,y}$) | Accuracy ($c_{18}$ \| $c_{J,x}$, $c_{J,y}$, $c_{J,z}$) | Accuracy ($c_{18}$ \| $c_{J,x}$, $c_{J,y}$) | Accuracy ($c_{28}$ \| $c_{J,x}$, $c_{J,y}$, $c_{J,z}$) | Accuracy ($c_{28}$ \| $c_{J,x}$, $c_{J,y}$) |
|---|---|---|---|---|---|---|---|
| Decision Tree | N | 79.2% | 66.3% | 79.6% | 75.2% | 79.2% | 66.1% |
|  | Y | 68.9% | 48.1% | 69.2% | 48.9% | 69.4% | 48.7% |
| Lin. Discr. | N | 80.0% | 71.9% | 80.2% | 76.2% | 80.0% | 71.9% |
|  | Y | 51.8% | 32.0% | 54.9% | 31.9% | 55.1% | 32.0% |
| Cubic SVM | N | 91.6% | 83.5% | 97.6% | 91.4% | 97.2% | 78.8% |
|  | Y | 57.0% | 15.9% | 51.9% | 10.5% | 53.6% | 13.8% |
| Fine k-NN | N | 94.4% | 79.8% | 93.4% | 87.6% | 92.7% | 74.7% |
|  | Y | 77.2% | 39.4% | 76.1% | 37.9% | 75.1% | 38.8% |
| Bagged Tree | N | 92.5% | 75.7% | 91.7% | 87.6% | 91.7% | 73.8% |
|  | Y | 76.7% | 39.7% | 76.1% | 38.9% | 74.6% | 39.3% |
| DNN (w=175) | N | 95.1% | 84.4% | 96.5% | 90.6% | 90.7% | 87.1% |

*B. Efficient Feature Space*

In the context of lower computation costs, and higher generalizability, previous work often accentuates the importance of effective feature extraction [65]–[67]. Within the next step of our research, we investigate how different input feature vectors influence the algorithms' accuracy. Following our method described in Figure 3, we tested how the respective accuracy of selected algorithms is affected when considering a subspace $C_i$ containing i relative body joint coordinates, where i = {9, 18, 28}. Moreover, we explored further possibilities for dimensionality reduction. In doing so, we compared the accuracy for a three-dimensional posture recognition ($c_{J,x}$, $c_{J,y}$, $c_{J,z}$) to a two-dimensional posture ($c_{J,x}$, $c_{J,y}$). After this, we repeated this procedure while integrating principal component analysis (PCA). PCA (explained variance $\sigma^2_{exp}$ = 95.00%) allows us to transform features and remove dimensions.

Without previous PCA, algorithms such as DNN and fine k-NN reach an accuracy of up to 96.50% and 94.40% respectively. This might indicate that these algorithms are suitable for HAR of AV. With further integration of PCA, however, accuracy of all algorithms considered dropped significantly. In general, a decrease of the accuracy of a model might not be a desirable outcome. Due to neglection of outliers in PCA, however, the models tend to be less overfitting to the LboroHAR dataset. Following this, these values give a more reliable estimation of the algorithmic performance when applied in realistic scenarios (Table V). On the other hand, outliers might be of non-neglectable importance for HAR in real-life applications. This is because an outlier can represent a person that has a different way of performing an activity than the majority of people considered (e.g. due to a disability, disease or drug influence). Our results show that an increase of the feature space does not lead to significant increase of accuracy. This observation is coherent with the curse of dimensionality discussed above.

The consideration of a 2-dimensional body posture representation leads to a significantly lower accuracy. Following this, the depth data of an RGB-D sensor is adding value for the classification procedure. We conclude that an efficient feature space can therefore be seen in ($c_9$ | $c_{J,x}$, $c_{J,y}$, $c_{J,z}$) or ($c_{18}$ | $c_{J,x}$, $c_{J,y}$, $c_{J,z}$). Within this feature space and with the integration of PCA, fine k-NN (77.20%) and bagged trees (79.70%) provided the highest accuracy for classifying HAR. The confusion matrix and receiver operating curve (ROC) can be seen in Figure 7 and Figure 8.

*C. Stationary and Dynamic Activities*

Our simulations indicate that there are significant differences in the classification of stationary (class 1-4) and dynamic (class 5-9) human activities. Regardless of the model applied, the classification of stationary classes reaches a comparatively higher accuracy (k-NN classifier, Figure 7; DNN Figure 8). For example, the confusion matrix in Figure 8 depicts that walking (class 5) and running (class 9) are often prone to misclassification, when feature vector is based on coordinates. In particular, for these two classes, the accuracy increases when using a velocity-based input vector (Figure 8). Due to the relative coordinate system, other distinctive characteristics (e.g. the so called "flight phase" in which none of the legs of the runner has contact with the ground) are of negligible importance.

While running (class 9) mostly got mistaken for walking (class 5), walking mostly got mistaken for walking while texting (class 6) and transition, or an object (class 7,8) (Figure 8). In all activities, the participant's arms are located at different distances and positions in front of their center of gravity. A distinction should therefore be possible, especially since the objects differ significantly in size and appearance. One reason for this misclassification can be seen in the selected feature space: no objects are included in the feature vector. We conclude that it might not be possible to realize a robust HAR without integrating the context of an activity into the feature space.

The classification of dynamic activities would potentially benefit from the integration of velocity into the feature vector.

However, given the curse of dimensionality, it would be useful to integrate only the speed of major body joints instead of simply adding together both feature spaces.

## VI. DISCUSSION

After presenting, describing, analyzing, and synthesizing our data, we will now discuss the generalizability and limitations of this work, and give an outlook for further improvements needed to realize the objective of AV in ambient-assisted living environments.

### A. Derivations from the Results

The findings of this work indicate that it is possible to realize HAR for autonomous vehicles in indoor environments using RGB-D sensors and cognitive approaches. Initial results show promising accuracy ($> 95.0\%$) for classifying human activities, in particular for stationary activities. However, considering the PCA, we found a significant performance decrease ($\ll 75.0\%$). It is expected that the accuracy under real conditions will deteriorate even more. For example, more realistic scenarios would include more noise into the input data, thus decreasing the probability of correct classification. The potential applications that build on our models interact in close physical contact with humans. Consequently, the accuracy of the individual classification models must be considered inadequate. A reliable statement of which algorithm is most suitable for our research goal can therefore be made exclusively by further experimental investigations of our models under real conditions. Thus, the present work should not be construed as an attempt to provide a fully functional solution. This research project is rather to be understood as a proof of concept that formed the foundation for which further research teams can utilize autonomous vehicles for the purposes of ambient-assistant living (e.g. smart wheel chairs).

Instead of relying exclusively on two-dimensional camera data, it seems useful to include the depth information for classifying activities. The relative coordinates of body postures provide a robust basis for classifying (stationary) activities. The inclusion of information about the speed of execution of an activity, however, may further improve the performance of the algorithms. Furthermore, our results suggest that only the use of a multimodal system of sensors and classification algorithms, which reduce individual inadequacies, can sufficiently realize HAR. This outcome coincides with ongoing efforts in computer vision as well as in activity recognition for AV in outdoor environments [21]. Although our experimental framework allows for exact reproducibility, it limits the generalizability of our research.

### B. Limitations of the current work
#### 1) Limitations of RGB-D Sensor

In addition to the applicability of different ML algorithms, we were interested in how the RGB-D sensor is suitable for our research purpose. For ambient assistant home applications, an AV (e.g. smart wheel chair) will most likely have its operative range between a human's feet and hips. Therefore, it is of particular importance that a human's lower body can be reliably recognized to prevent accidents in indoor environments. During our experiment, however, we found that black fabrics, especially sweatpants, limit the correct recognition of human silhouettes. We therefore conclude that the RGB-D sensor alone cannot meet the requirements of a robust HAR system.

#### 2) Stationary Conditions

During our data collection, the sensor testbed was in a stationary state. This was due to two reasons. In order to ensure comparability of different sensors, the activities needed to be carried out exclusively in the optimal operating range of all available sensors. The RGB-D used was developed for gaming applications and it is therefore to be assumed that its performance is better if the sensor is stationary during data collection. On the other hand, we were interested in the general sensor-algorithm applicability, rather than their respective robustness against external influences and noise such as vibrations of the vehicle. Our findings are therefore limited to scenarios in which an AV classifies human activities from a stationary perspective. This includes, for example, checking the environment before driving off.

The tracking of 16 participants started from a standardized position (T-position). This setup results in two further limitations. Although we included more than 7000 postures in our models, conclusions about the accuracy of the algorithm for a larger quantity are critical due to the relatively small number of participants. For example, one could address this issue of a small dataset by creating an artificial dataset containing data that is added with gaussian noise. With this small number of participants, the individual characteristics of the execution of an activity have a higher weight, resulting in the risk for a biased model. The approach of an artificial dataset would further amplify this inherent distortion. Additionally, people do not take a standard position before performing a new activity. Consequently, an algorithm operating under real conditions must be able to track a person without a prior "T-position". Also, our approach is not intended to classify activities of persons who are in a group.

The classification was realized in retrospect with a pre-defined feature space. We have not investigated how the trained models perform in real-time scenarios, yet. For example, the delay of the classification could be significantly different. Moreover, our findings regarding an efficient feature space suggest that relative coordinates should be combined with other information (e.g. velocity). Because distinguishing factors may differ for individual activities, previous work often recommended a dynamic feature extraction instead of a pre-defined setting [46], [68]. Although the improvement of feature extraction was not a focus of this work, a dynamic feature extraction might have had an impact on an algorithm's performance.

#### 3) Spatial Time Invariant Data Representation

An activity is often defined as a type of human behavior in relation to his/her environment [4]. Following this, a robust HAR system needs to include a contextual evaluation of the activity and its relation to the environment. The input data for the ML algorithms, however, is represented as a number of frames containing data of body postures. This implies that classification of a human activity is not possible, when the distinguishing factor of an activity is the interaction with the environment. For example, our approach cannot accurately distinguish class 6 (human texting while walking) from class 7

(carrying objects) because the underlying feature vector does not contain any information about the smart phone or the carried object. Therefore, the models developed in this work are only suitable for activities that can be sufficiently characterized by the posture alone (e.g. sitting and standing).

*C. Generalizability of the ML algorithms*

A central challenge in ML is how to ensure an algorithm will perform well not just on the training data but also on new inputs [69]. However, a defining characteristic of an ML classifier is that the mathematical model which predicts the class of a new input value is developed internally [70]. In general, these models are incomprehensible to a human supervisor. As a result, we can only approximately quantify how the selected algorithms will perform in reality. Nonetheless, it is possible to make statements about the basic behavior of the underlying models.

As the chosen SVM is using a cubic kernel function for classification, this approach generally implies a comparatively higher risk of overfitting. Moreover, SVM represents decisions rather than posterior probabilities [62], [71]. In the context of dynamic feature extraction, an SVM might also be problematic, since adjustment in the underlying dimension requires a retraining of the model. A possible alternative might be seen in the relevance vector machine (RVM). This classification (or regression) technique is based on a Bayesian sparse kernel that shares many of the characteristics of an SVM, whilst avoiding its limitations discussed above [62]. RVM provides posterior probabilistic outputs and has much sparser solutions than the SVM. A major disadvantage of RVM is the relatively long training time.

Decision trees are also prone to overfitting. This especially applies to tree classifiers with a certain depth. To limit overfitting without substantially increasing errors caused by bias or high variance, it might have been beneficial to consider random forests instead of fine decision trees, and bagged trees. This method describes a set of multiple decision tress whose results are aggregated into one final result.

Overall, it can be stated that due to the characteristics of the ML algorithms, no explicit conclusion about generalizability can be drawn, unless the models were experimentally tested in other scenarios. This directly points towards suggestions for further research.

*D. Future Work*

This work explores how ML algorithms can enable HAR for AV in indoor environments. In order to derive reliable driving policies for AV in indoor environments, however, it is necessary to conduct further work. We introduce six major aspects that would mitigate the limitations of this study and improve the algorithmic models, thus further narrowing down the existing gaps in the research field of interest: (a) algorithm-sensor performance, (b) sensor fusion, (c) enhanced feature extraction, (c) dynamic data capture, (d) real-time applicability and prediction of intentions, (e) human-object recognition.

Here we only consider input data captured by an RGB-D sensor. It would be of great interest to see how the performance of the selected algorithms changes depending on different sensor inputs. Thus, one objective is to find out which interaction of sensor and algorithm provides the best performance for HAR. Since the activities were recorded by six sensors, the LboroHAR dataset provides a basis for first strides towards an optimal match of sensors and algorithms.

After identifying the best algorithm-sensor match, a logical step would be to fuse multiple data streams for a robust multi-modal HAR system. The sensor fusion implied herein could help mitigate the observed weaknesses of individual sensors.

In *section 3* and *4* we also emphasized the importance of efficient feature extraction. Following the approach of [72], we decided to classify the activities using single body postures. However, it might be beneficial to the classification accuracy if HAR is performed based on a sequence of frames $n > 1$. For example, one could think of classifying a contiguous sequence of postures to eventually predict the class of the "parent activity". Another possibility would be to integrate multiple postures (represented by body joints only) into a single Vector $F$ feature. Here it would be of the utmost importance that the dimensionality of $F$ does not increase significantly (curse of dimensionality). However, both approaches require a larger amount of data because the available training and test samples decrease at least by the factor $\frac{1}{n}$. Regardless of the chosen method, we consider it important to further optimize feature extraction. In [14] a recent work addresses this topic. Here, various features are aggregated into a feature vector of high dimension. However, the crucial vector optimization that counteracts the curse of dimensionality is still missing.

Input data captured by a moving sensor testbed will most certainly affect the accuracy as well, since it adds additional noise. Thus, a future step in the research agenda might be seen in a repetition of the experiment with a dynamic and realistic framework.

As discussed, a major limitation of our approach is the focus on body joints. For more complex activities, the environment and the object that is used or altered during an activity needs to be considered as well. The objective of enabling the recognition of humans with objects can also be realized with our dataset, the visual perception of objects of interest is given by the camera, while depth data is provided by point clouds.

## VII. CONCLUSIONS

The availability of appropriate datasets is a crucial element of any data science activity. Human activity recognition is not any different to this. While most of the HAR activity has focused on a single sensing modality, this work discussed the utility of multimodality for robustness of HAR. Furthermore, a specific limitation of multi-modal datasets for indoor environments was identified and in response we focused on indoor HAR for applications such as autonomous vehicles, and healthcare assistant robots. We comprehensively the existing datasets and identified the experimental setup needed to capture the open dataset proposed in this paper. To make the dataset useful to as many researchers as possible, who work on different branches of machine learning algorithms (i.e. other than deep networks), we developed a novel data-preprocessing technique for dynamic feature extraction from the body pose detections. This algorithm allows for a flexible, pre-determined selection of body joints. We analyzed a selection of 6 different ML algorithms on RGB-D sensor modalities: decision trees, linear discriminant, cubic SVM, fine k-NN, bagged tree, and DNN. To justify the proposed preprocessing algorithm we assess how various input parameter influence the performance of the selected ML algorithms. Our investigations indicate that activities can be successfully recognized by the pre-processing method proceeded by state-of-the-art ML algorithms. Most accurate results, for RGB-D modality, can be reached when using the three-dimensional body joints as representation of postures with relative coordinates as input feature (positive likelihood ratio with DNN: 99%). While enhancing our feature extraction we observe the curse of dimensionality, the principle of overfitting and the bias-variance trade-off. While HAR can be realized by relying only on RGB-D data, we argue that only the use of a multimodal system of sensors and classification algorithms can sufficiently realize HAR for safety critical systems such as autonomous vehicles to amplify individual human capabilities in ambient-assistant living. It is expected that the proposed dataset and algorithms for pre-processing will advance the application of multimodal machine learning techniques for human activity recognition.

Biographies

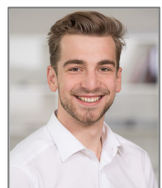

**Mirco Möncks** is currently reading for an MSc in Industrial Engineering and Business Management at Karlsruhe Institute of Technology. Mirco holds an MSc in Internet Technologies with Business Management from Loughborough University London. He received awards from Loughborough University London for academic excellence, and for the most impactful postgraduate dissertation project of the year 2018. Before that, he was working as a bachelor's thesis student and intern at the production system of AUDI AG. In this role Mirco was improving the companies' logistics processes, and solving problems occurring at the assembly lines. In 2017 Mirco graduated from Ilmenau University of Technology with a BSc in Industrial Engineering and Management with a focus on mechanical engineering. His current research interests are in human activity recognition for ambient-assistant living, and wearable technologies for development of manufacturing skills in human-centered production systems.

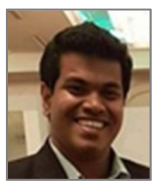

**Varuna De Silva** received his Ph.D. from University of Surrey in United Kingdom in 2011. He was a post-doctoral researcher in the same institute between 2011-2013, working on video and image processing for Multiview video broadcasting applications. Since 2013 November, he worked as a senior algorithms developer for image signal processors at Apical Ltd, UK (Now part of ARM Plc). In April 2016, he joined Loughborough University, as a lecturer in digital engineering. He is the recipient of IEEE Chester Sall award from the Consumer electronics society in 2011, and Vice-Chancellors award for the best Post graduate research student of the year 2011. His current research interests include sensor data processing and artificial intelligence for driverless vehicles and applied data science.

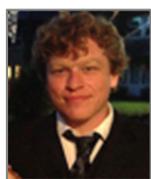

**Jamie Roche** is currently reading for a PhD in Intelligent Mobility after having undertaken an M.Sc. in Mobile Internet at Loughborough University London. Prior to coming to London Jamie worked as a Consulting Forensic Engineer for seven years with Denis Wood & Associates in Dublin Ireland. In 2008 Jamie graduated from Trinity Collage Dublin with a M.Sc. in Bioengineering. Prior to undertaking his Bioengineering degree Jamie worked as a project engineer with Irish Rail, a Water Sanitation and Hygiene (WASH) Engineer with Help International (NGO) and a graduate Engineer at Schering Plough Pharmaceuticals. In 2002 Jamie graduated from Dublin City University with a

B.Eng. in Mechatronics. His current research interests are in cognitive approaches to multimodal sensor data perception.

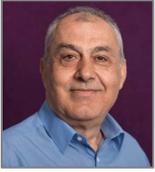

**Ahmet Kondoz** was awarded his PhD in Signal Processing and Communication from the University of Surrey in 1987 where he was employed as Lecturer and Reader in 1988 and 1995 respectively. He was promoted to Professor in Multimedia Communication Systems in 1996. Ahmet took part in setting up of the world renowned Centers CCSR and the I-Lab at the University of Surrey before joining Loughborough University London as the Director for the Institute for Digital Technologies. Ahmet has been involved in the coordination of large national and International research projects funded by the European Commission. He has been collaborating with major EU Industries and Universities. He coordinated FP6 VISNET, FP7 DIOMEDES and ROMEO projects. He contributed as a core partner to SUIT, NEWCOME and MUSCADE. Currently he is the coordinator of CLOUDSCREENS.